\documentclass{template/iosart2c}

\usepackage[T1]{fontenc}
\usepackage{times}%
\usepackage[square,sort,comma,numbers]{natbib}
\usepackage{amsmath}
\usepackage{dcolumn}
\usepackage{graphics}

\usepackage[super]{nth}
\usepackage{hyperref}
\usepackage{graphicx}
\usepackage{caption}
\usepackage{algorithm}
\usepackage{algpseudocode}
\usepackage{subcaption}
\usepackage{comment}
\usepackage{inconsolata}
\usepackage{booktabs}
\usepackage{listings}
\usepackage[frozencache, cachedir=minted-cache]{minted}
\usepackage{afterpage}
\usepackage{lipsum}
\usepackage{lscape}

\newcolumntype{d}[1]{D{.}{.}{#1}}

\firstpage{1} \lastpage{5} \volume{1} \pubyear{2023}

\begin{document}
\begin{frontmatter}                           

%
\begin{center}
\title{LuckyMera: a Modular AI Framework for Building Hybrid NetHack Agents}
\end{center}

\runningtitle{LuckyMera: a Modular AI Framework for Building Hybrid NetHack Agents}

\author[A]{\fnms{Luigi} \snm{Quarantiello}\thanks{l.quarantiello@studenti.unipi.it}},
\author[A]{\fnms{Simone} \snm{Marzeddu}},
\author[A]{\fnms{Antonio} \snm{Guzzi}}
and
\author[A]{\fnms{Vincenzo} \snm{Lomonaco}}
\runningauthor{L. Quarantiello et al.}
\address[A]{Department of Computer Science, University of Pisa}

\begin{abstract}
In the last few decades we have witnessed a significant development in Artificial Intelligence (AI) thanks to the availability of a variety of testbeds, mostly based on simulated environments and video games.
Among those, roguelike games offer a very good trade-off in terms of complexity of the environment and computational costs, which makes them perfectly suited to test AI agents generalization capabilities.
In this work, we present \textit{LuckyMera}, a flexible, modular, extensible and configurable AI framework built around NetHack, a popular terminal-based, single-player roguelike video game. 
This library is aimed at simplifying and speeding up the development of AI agents capable of successfully playing the game and offering a high-level interface for designing game strategies.
LuckyMera comes with a set of off-the-shelf symbolic and neural modules (called \emph{"skills"}): these modules can be either hard-coded behaviors, or neural Reinforcement Learning approaches, with the possibility of creating compositional hybrid solutions. Additionally, LuckyMera comes with a set of utility features to save its experiences in the form of trajectories for further analysis and to use them as datasets to train neural modules, with a direct interface to the \textit{NetHack Learning Environment} and \textit{MiniHack}.
Through an empirical evaluation we validate our skills implementation and propose a strong baseline agent that can reach state-of-the-art performances in the complete NetHack game. LuckyMera is open-source and available at \url{https://github.com/Pervasive-AI-Lab/LuckyMera}.
\end{abstract}


\end{frontmatter}

\section{Introduction}

In the last few years, Artificial Intelligence algorithms achieved astonishing results in a wide range of tasks, exploiting both classical, symbolic approaches and data-driven methodologies from the field of Machine Learning \cite{deep_learning}.
Research in this area was sustained and encouraged by the availability of several benchmarks, needed to experiment with new architectures and technologies.
Video games in particular, gained a great popularity since they offer challenging experiences, similar to real-world problems, at a much smaller cost; therefore, they are excellent playgrounds to study and test different approaches.
AI was applied to different environments with surprising success, \emph{e.g.} winning against the world chess champion \cite{deep_blue} and, especially with the introduction of deep architectures in the field of Reinforcement Learning (RL), beating professional players in Go \cite{go_rl} and Dota 2 \cite{dota2_rl}.

Among the wide variety of games available, \emph{roguelikes} are of particular interest due to their unique features.
Roguelikes\footnote{\url{http://www.roguebasin.com/index.php/Berlin\_Interpretation}} are turn-based, role-playing games, with a deep focus on cautious exploration, fighting enemies and wise resource management. This kind of game also features random generation of the levels structure, together with the type of enemies and objects the player will find, and \emph{permadeath}, meaning that there is no checkpoint during the game, and players have to start from the first level each time they die.
Because of these features, roguelike video games are extremely challenging, since players are required to deal with a variety of situations, each time having to overcome many levels, where a single mistake could ruin an entire run.
Given these characteristics, roguelike video games are perfectly suited to test AI agents ability to generalize in increasingly complex environments.

\begin{figure}
    \centering
    \includegraphics[width=0.95\linewidth]{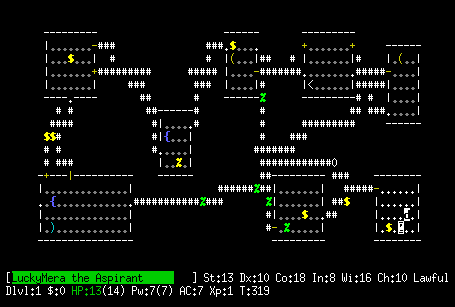}
    \caption{An example of the NetHack ASCII interface.}
    \label{nethack_floor}
\end{figure}

\emph{NetHack}, published in 1987, is one of the earliest and most popular roguelikes; here, the player controls the hero, selected among different races, roles and alignments, with the objective to retrieve the \emph{Amulet of Yendor} by exploring over 50 random generated floors, and offer it to a deity to became a \emph{demigod}.
Each level is made of rooms connected by corridors, filled with monsters, objects and other peculiar features, such as shops or altars.
NetHack, in its original version, provides a simple terminal interface (Fig. \ref{nethack_floor}), depicting the map of the current level.
In addition, it shows a message on top of the screen offering additional information, and a bottom line with character statistics.

The game offers a complex, procedurally generated open world with sparse rewards, forcing the agent to explore, reason and acquire knowledge about hundreds of entities.
\emph{NetHack is considered one of the most difficult games for humans}, and a hard challenge for modern RL models as well; in fact, current best models are only comparable to human beginners \footnote{in \url{https://nethackwiki.com/wiki/Beginner}, a beginner score is less than 2,000 score points}.

In this work, our objective is to present a complete and integrated framework, that can facilitate the research in AI exploiting the enviroment offered by NetHack. We argue that our framework is an effective tool to design a number of agents playing the game, using both classical symbolic AI solutions and Machine Learning ones.
To the best of our knowledge, \emph{this is the first open-source framework aimed at the definition of AI agents built around the game of NetHack and the NetHack Learning Environment}.

Our main contributions can be summarized as follows:

\begin{itemize}
    \item We introduce LuckyMera\footnotemark, a modular and extensible framework for building intelligent agents for NetHack. It integrates different AI paradigms, \emph{i.e.} symbolic and neural approaches, and offers the possibility to easily define custom modules to solve specific tasks;
    \footnotetext{Being a modular framework, it is similar to a \emph{chimera}: a mythological hybrid creature composed of different animal parts. But NetHack is difficult, so it needs to be lucky \emph{"mera"} (which stands for \emph{"a lot"} in Sardinian)!}
    \item We discuss different approaches to the game, in particular Imitation Learning, Reinforcement Learning and Neuro-Symbolic methods. We perform ablation studies concerning these components, to analyze their performance;
    \item We show how a bot build with LuckyMera is able to reach state-of-the-art performances within the top 6 bots of the NeurIPS NetHack Challenge 2021 among over 600 submissions \cite{nle_challenge}.
\end{itemize}

\begin{figure}[t]
\centering
\includegraphics[width=\linewidth]{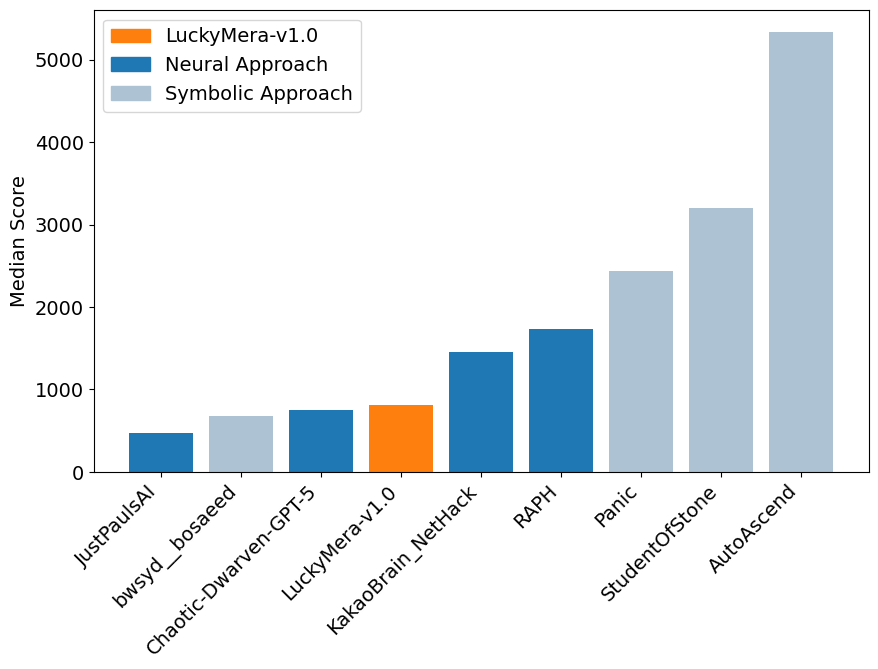}
\vspace{-0.5cm}
\caption{Results obtained by a LuckyMera agent, compared with the state-of-the-art bots from the NeurIPS NetHack Challenge 2021.}
\label{fig:results_challenge}
\end{figure}

\section{Related Work}
In this section, we review some studies related to our work, starting from the virtual environments defined around the game of NetHack for AI agents, and relevant AI approaches, namely Rule-Based, Imitation Learning and Reinforcement Learning methodologies.
\subsection{NetHack as AI Testbed}
As we will see more in depth in Section \ref{sec:symbolic_bots}, the environment offered by NetHack was widely used to develop and test intelligent agent capable of playing the game.  
An interesting approach is the one proposed in \cite{nethack_occupancy}, in which the authors present a solution to explore the levels of the game exploiting the concept of occupancy maps, especially popular in robotics.
In RL research, the first example of usage of NetHack is gym\_nethack \cite{gym_nethack1, gym_nethack2}, which offers an interface to the game through a Gym \cite{openai_gym} environment.
However, in this case, the dynamics were heavily modified by removing several obstacles, resulting in a much simpler version of the game.

In this work, instead, we make deep use of the \emph{NetHack Learning Environment} (NLE) \cite{nle}: a Gym environment that leaves the game mechanisms unchanged.
It is designed to wrap the entire game, returning all the observation available from the game, \emph{i.e.} the map of the level, the current agent's statistics, the textual message showed to the user and information about the inventory.
The environment has 93 available actions, divided in 16 movement actions and 77 command actions.
NLE is of particularly interesting because it is able to combine a complex environment with a fast simulator, being extremely efficient and computationally lightweight.
Since current architectures cannot win the game, MiniHack \cite{minihack} was released: a framework defined on top of NLE, which proposes a set of simpler environments, together with the possibility to easily design new tasks. 
The tasks proposed can be mainly divided into navigation tasks, in which the agent has to reach a goal position, and skill tasks, which involve more complex abilities, such as using potions, selecting the appropriate armor and fighting more powerful monsters. 

We were deeply inspired by the results obtained in the NeurIPS 2021 NetHack Challenge \cite{nle_challenge}. Our framework was developed using mainly the \texttt{challenge} task, but it is completely environment-independent, and it work well with all the tasks proposed in NLE and MiniHack.

\subsection{First AI Bots for NetHack} \label{sec:symbolic_bots}
From its initial release, there have been several bots addressing the problem of NetHack.
One of the first able to achieve significant results is TAEB \footnote{\url{http://taeb.github.io/index.html}}, a modular framework for designing automatic and semi-automatic players. It uses the publish/subscribe paradigm to perform the communication among the different components; for the pathfinding task, it employs Dijkstra's algorithm \cite{dijkstra}.
The first symbolic bot able to "ascend", \emph{i.e.} win the game, was BotHack \footnote{\url{https://github.com/krajj7/BotHack}}. 
Its architecture is particularly noticeable for the accurate recognition of the kind of floor the agent is exploring, and the use of the A* algorithm \cite{a*} for the navigation tasks.
Nonetheless, it was able to achieve these results mostly using an exploit present in older NetHack version, which is no longer applicable in the current game.
The current best open-source NetHack bot is AutoAscend \footnote{\url{https://github.com/maciej-sypetkowski/autoascend}}, winner of the NeurIPS 2021 NetHack Challenge. It implements a set of high-level strategies, each handling a specific behavior and wrapping multiple actions, and selects one based on its priority.

Although these approaches are able to obtain good results at the game, none of them offers a valid research platform, as we do with LuckyMera.
Their goal was to create performance-oriented agents to win the game, while our main objective is to provide a development-oriented framework, to train, integrate and test neuro-symbolic approaches. 

\subsection{Imitation Learning Approaches}
Imitation Learning is a Machine Learning technique in which the agent, to learn an intelligent behavior, instead of relying on the interaction with the environment, is provided with a set of demonstrations from an expert \cite{imitation_learning}.
The agent's objective is to mimic the expert's actions, hopefully achieving an optimal policy, following a form of Supervised Learning.
The dataset contains trajectories of experiences, made of \emph{state-action} pairs; in particular, the trajectories will be in the form of
$$\tau = \{(s_0^*,a_0^*),(s_1^*,a_1^*),\dots, (s_n^*,a_n^*)\}.$$
It is critical to notice that, when performing Imitation Learning, the agent should not copy the expert's behavior unconditionally; instead, it should extract key information from the trajectories, being able to generalize and achieve good performance also in states never seen before.

One of the simplest Imitation Learning algorithms is Behavioral Cloning (BC): given a \emph{state-action} pair $(s_t, a_t^*)$, the objective is to learn a policy $\pi$ by minimizing a loss function $L(a_t^*, \pi(s))$, assuming the pairs are \emph{i.i.d.}.
BC has been shown to achieve good results especially in environments with relatively small state space, so that it can be covered for the most part by the expert's demonstrations, e.g. autonomous driving \cite{alvinn}.
Nonetheless, in most cases it can be quite problematic due to the \emph{i.i.d.} assumption.
An improvement of BC is DAgger \cite{dagger}, that employs an iterative process in which it first performs Supervised Learning to learn a policy, like in BC. It then uses it to produce observations, queries the expert on those observations, and integrates the dataset with these new demonstrations.
A different approach to Imitation Learning is represented by Inverse Reinforcement Learning \cite{irlsurvey}, in which the idea is to learn the reward function by observing the demonstrations from the expert, and then use it to find the optimal policy with Reinforcement Learning algorithms.
NetHack is particularly convenient to perform Imitation Learning, thanks to the NetHack Learning Dataset \cite{nld}. It collects both state transitions from human games, and state-action trajectories generated by the winner of the NetHack Challenge 2021, AutoAscend.

We will review in detail the implementation of the Behavioral Cloning algorithm we offer in LuckyMera in Section \ref{sec:training_neural_agents}.

\subsection{Reinforcement Learning Approaches}
Reinforcement Learning algorithms are usually tested in simulated environments, like games.
RL approaches have shown superhuman capabilities in classical games, such as Go \cite{go_rl} and Chess \cite{chess_rl}; furthermore, there were also works on more complex, multiplayer games, like StarCraft II \cite{starcraft_rl} and Dota 2 \cite{dota2_rl}.
Several studies on Reinforcement Learning were conducted on the NetHack environment.
The MiniHack suite was used to test the E3B algorithm \cite{e3b}, a method for defining intrinsic exploration bonuses based on learned embeddings of previous states.
Chester et al. \cite{oracle-sage} proposed a hybrid approach, using symbolic planning for low-level actions, and Reinforcement Learning to train a meta-controller; they show that this method surpasses the baseline algorithms in a custom MiniHack environment.
Powers et al. \cite{cora} presented CORA, a platform for Continual Reinforcement Learning, providing MiniHack as one of the benchmark environments.
Using the MiniHack tasks from CORA, Kessler et al. \cite{latent_world_models} studied a task-agnostic, model-based method for Continual Reinforcement Learning, showing it to be a strong baseline compared to state-of-the-art approaches.

LuckyMera is instead an approach-agnostic research platform; it can be expanded to test any Artificial Intelligence method or paradigm, including Reinforcement Learning ones. It is particularly convenient because of the possibility to train targeted skills, tackling a more feasible problem, and then to integrate them with the other modules offered by the framework.

\section{The LuckyMera Framework}
LuckyMera is a framework for simplifying the development of Artificial Intelligent agents able to play the game of NetHack, designed following the principles of modularity, extensibility and configurability.
The main objective of the architecture is to provide a high-level interface for defining game strategies, represented in the code through the \texttt{Skill} abstraction.
A skill is defined as \emph{any complex activity - a composition of several elementary actions to achieve a given goal - that can be planned and executed in a given state of the NetHack game}.
Each skill is defined as a separate module, so that it can communicate with the main components of the framework without limitation on the implementation details. In the following sections, we will present some examples of modules released with the framework.
A complete overview on how to use the framework can be found in Appendix \ref{app:usage}.

\begin{figure}
    \centering
    \includegraphics[width=\linewidth]{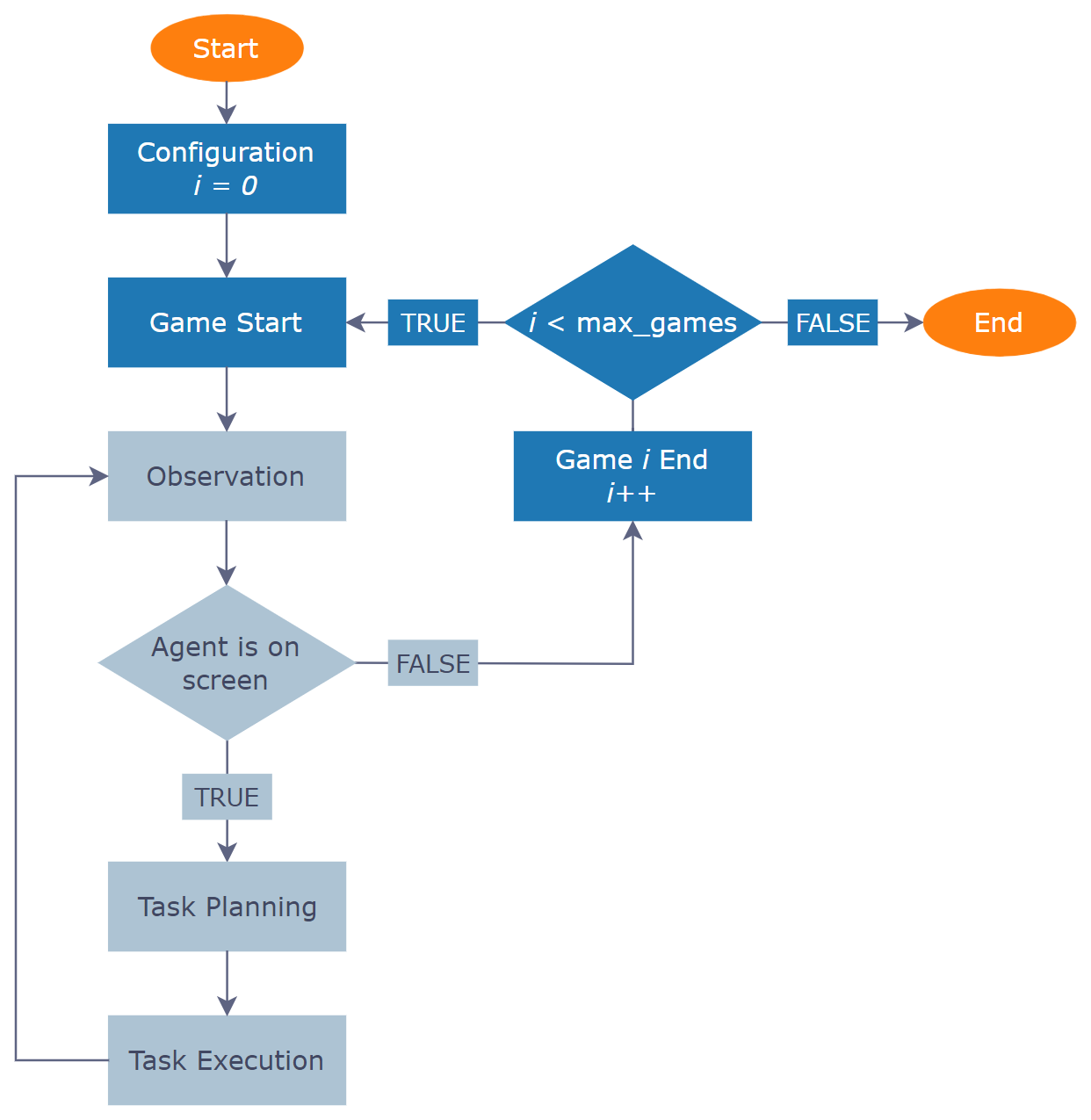}
    \caption{Flowchart of the behavior of the LuckyMera agent. Essentially, it iterates through planning and execution. During skill planning, it gets the highest priority skill that can be planned; then, this skill is actually executed.}
    \label{fig:jb3_flow}
\end{figure}

\subsection{Design of the Agent}
The LuckyMera framework is released together with the implementation of an AI agent, designed following the modular structure, that represents a useful baseline for further studies.
At the highest level of abstraction, the agent's strategy simply consists of the iterative execution of the highest-priority plannable skill in the current state; the user can easily set the priority of each skill via a configuration file.
In Figure \ref{fig:jb3_flow} there is a high-level representation of the system execution flow.

The agent's main execution flow is made by the iteration, throughout the duration of the game, of the two principal phases: \emph{skill-planning} and \emph{skill-execution}.
The planning of a skill begins with the analysis of the game state, derived from the NLE observations.
During this phase, the agent verifies if the skill can actually be executed, \emph{i.e.} its preconditions are satisfied, and performs some preliminary steps, depending on the nature of the skill itself.
The module implementing a skill should provide the planning method, in compliance with the framework's main component interface.
Thanks to this abstraction, the system is perfectly compatible out-of-the-box with any symbolic and neural skill implementation, and in general with any AI module; we will discuss this feature more in detail in Section \ref{sec:skills_integration}.
Once the planning of a skill succeeds, the agent performs the skill-execution phase.
Each skill provides the implementation of a series of actions needed to perform its plan. 

The framework offers also the possibility to define custom modules, by extending one of the classes representing the skill concept.
\texttt{Skill} is the base class for defining new skills; it is an abstract class, therefore it provides only some general-purpose methods, without a real implementation of planning and execution. In most cases, custom modules should inherit from this class. 
Another example is \texttt{ReachSkill}, which concerns simple navigation skills, offering an execution method for reaching specific locations in the game world.
Similarly, the \texttt{HiddenSkill} class implements useful methods for finding secret passages or areas in the game.

In addition to the high-level modules, the architecture also provides low-level solutions for interacting with the game.
In particular, it is based on the \texttt{GameWhisperer} class, which deals with the interaction with the NLE framework, offering several refinements to the low-level observations; it also defines methods to encapsulate multiple NetHack commands into single, more abstract atomic commands.
On the other hand, to handle the navigation in the game world, the system leverages on the \texttt{DungeonWalker} class. It offers some useful functionalities for pathfinding and exploration. To do so, it employs the A* algorithm \cite{a*}, using the octile distance heuristic \cite{octile_distance} to take into account also diagonal movements.

\subsection{Features of the Framework}
Here, we will discuss some of the main features that LuckyMera offers, which make it an integrated framework for quick testing of new approaches, automatic creation of labeled trajectories and training of Machine Learning models. 

\subsubsection{Skills Integration} \label{sec:skills_integration}
\begin{figure}
    \centering
    \begin{subfigure}{0.49\linewidth}
        \centering
        \includegraphics[height=3.29cm]{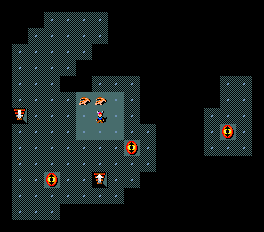}
        \caption{\texttt{Room-Ultimate-15x15}}
    \end{subfigure}
    \begin{subfigure}{0.49\linewidth}
        \centering
        \includegraphics[height=3.29cm]{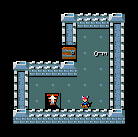}
        \caption{\texttt{KeyRoom-S5}}
    \end{subfigure}
    \caption{MiniHack environments used for the training of the RL agent}
    \label{fig:minihack_envs}
\end{figure}

The main component of the framework offers a simple, straightforward interface, making it easy to integrate skill modules.
Each module should inherit from one of the classes representing the \texttt{Skill} concept and, by doing so, define its own planning and execution methods.
All the actions that a LuckyMera agent can perform are defined as skills, and are executed following a priority list.
The strategy the agent follows is determined by the priority assigned to each module. In the framework, the order of the modules can be easily changed through a configuration file.
In our tests, we let the agent adopt a \emph{cautious} strategy, in order to maximize the score\footnote{\url{https://nethackwiki.com/wiki/Score}}.
In particular, the top-priority actions are the ones that can help the bot overcoming dangerous situations, like praying to receive resources, engraving the name ``\emph{Elbereth}'' to scare enemies and run.
After those actions, the agents checks if it can fight nearby monsters.
Otherwise, if it is in a safe circumstance, it can explore the unseen parts of the dungeon, or search for hidden rooms and corridors.
The complete list of the currently implemented skills is available in Appendix \ref{app:skills}. 
It includes both the symbolic skills and the ones coming from the integration of the neural modules.

Besides these skills, it is possible to integrate any external module compliant with the interface.
In fact, the framework allows for the import of any given model, so that new actions --- or new strategies for already defined operations --- can be implemented, or trained neural models can be used to perform specific tasks.
As an example, the framework was tested with the integration of a neural RL agent, trained using the IMPALA algorithm \cite{impala}, with the implementation from TorchBeast \cite{torchbeast}. Since the task of playing the entire game of NetHack is too difficult for current RL approaches, the training was executed on some MiniHack environments, which offers more controlled and feasible challenges.
The environments selected are \texttt{Room-Ultimate-15x15} and \texttt{KeyRoom-S5}, represented in Figure \ref{fig:minihack_envs}. In this case, the game map is depicted using the \texttt{pixel} observation from MiniHack, which is offered in addition to the standard ASCII interface.
To improve the performance of the pure neural agents, the trained models were then integrated with some prior knowledge about the game, in the form of a set of simple, generic rules.
More details about the symbolic rules are available in Section \ref{rl_results}.
\subsubsection{Trajectory Saving}
\begin{figure}
  \begin{subfigure}[b]{\columnwidth}
    \centering
    \includegraphics[width=\linewidth]{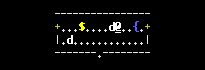}
  \end{subfigure}
  \vfill 
  \vspace{2mm}
  \begin{subfigure}[b]{\columnwidth}
    \centering
    \frame{\includegraphics[width=\linewidth]{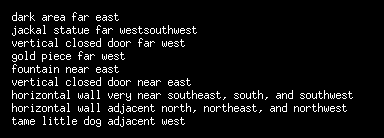}}
  \end{subfigure}
  \caption{One level of NetHack viewed with different representations. The top image is the standard ASCII interface, while the bottom image shows the language representation provided by the \texttt{nle\_language\_wrapper}}
  \label{fig:map_and_text}
\end{figure}

The architecture we propose comes with the possibility to save the experiences of the bot with the environment, in the form of trajectories of \emph{state-action} pairs.
Given the performance of a LuckyMera bot, its behavior is meaningful and valid in the context of the game, and it could be used as an expert, \emph{e.g.} in Imitation Learning applications.
Within the framework, it is possible to exploit the capabilities of the bot to define a dataset of experiences, which is automatically labeled with the actual action performed by the agent.
The trajectory saving mechanism is independent of the type of observation: it is possible to store any element present in the observation space defined by NLE, by defining them at runtime.
The framework integrates also the \texttt{nle\_language\_wrapper}\footnote{\url{https://github.com/Pervasive-AI-Lab/nle-language-wrapper}}, which translates the non-language observations from NetHack, \emph{e.g.} \texttt{glyphs} and \texttt{chars}, into corresponding language representations (Figure \ref{fig:map_and_text}).
These language observations can be selected in the saving process as well, and could be useful in the fine-tuning of language models.
\subsubsection{Training of Neural Agents} \label{sec:training_neural_agents}
The LuckyMera framework offers also the possibility to perform training processes on the NetHack environment, providing an interface that can handle any training algorithm.
In fact, the architecture comes with an abstract class that represents a generic training procedure, that should be extended to define a specific algorithm.
In this way, the training process is strictly incorporated in the system, so that it is easy to evaluate the model performance and integrate it with the other modules, giving the possibility to also define hybrid architectures.
As an example, we provide the implementation of the Behavioral Cloning algorithm, one of the most intuitive approaches for Imitation Learning; it is briefly described in Algorithm \ref{alg:BC}.

\begin{algorithm}
\caption{\textit{Behavioral Cloning}}\label{alg:BC}
\begin{algorithmic}
\Ensure a \textit{policy} $\pi_{\theta}$ trained on the problem 
\While{$L(a^*,\pi(s))$ is not small enough}
    \State Collect trajectories $\tau_1, \dots, \tau_n$ from the expert.
    \State Get all the $(s_i^*, a_i^*)$ from each $\tau_i$, as \emph{i.i.d.} pairs.
    \State Learn \textit{policy} $\pi^*$ by minimizing $L(a^*,\pi(s))$.
\EndWhile
\end{algorithmic}
\end{algorithm}

\begin{figure*}[t!]
\begin{subfigure}{0.40\textwidth}
    \centering
    \includegraphics[height=4.6cm]{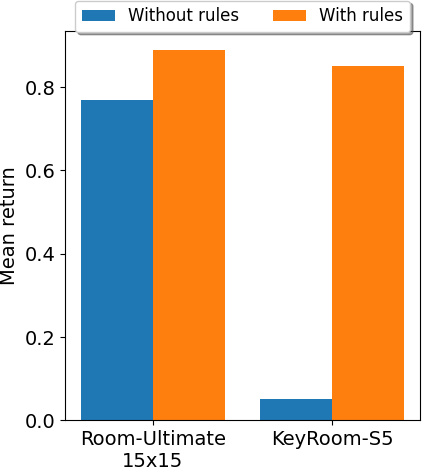}
    \captionsetup{justification=centering}
    \caption{Results of the RL agent in the two environments, analysing the impact of the rules integration}
    \label{fig:rules_integration}
\end{subfigure}
\hspace{1cm}
\begin{subfigure}{0.40\textwidth}
    \centering
    \includegraphics[height=4.6cm]{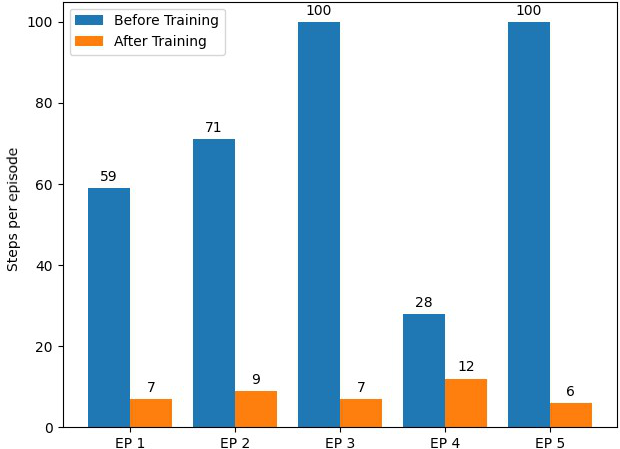}
    \captionsetup{justification=centering}
    \caption{Results obtained in five different episodes. In each of them, the agent had a maximum of 99 steps to reach the goal.}
    \label{fig:bc_results}
\end{subfigure}
\caption{Results of the ablation studies performed on the framework}
\end{figure*}

\section{Empirical Evaluation} \label{sec:results}
In this section, we evaluate the results of the experiments we conducted with the LuckyMera framework.
Firstly, we will analyse the ablation studies we performed on some modules of the architecture, namely the hybrid RL module and the Imitation Learning approach, to see their individual performance.
We will then present the results obtained by the LuckyMera agent we release, comparing it with the participants in the NeurIPS 2021 NetHack Challenge.

\subsection{Reinforcement Learning Approach Results} \label{rl_results}
We integrated a Reinforcement Learning approach, based on the IMPALA algorithm.
As testing environments, we used the \texttt{Room-Ultimate-15x15} and \texttt{KeyRoom-S5} tasks from MiniHack.
Initially, we tested the pure neural approach, then we integrated it with some prior knowledge about the problem, in the form of basic symbolic rules, expressed using first-order logic.
Those rules were used to increase the probability of crucial actions, like attacking nearby monsters and moving towards the key.
In particular, the rules we employed are showed in Table \ref{tab:fol_rules}.

\begin{table*}
\caption{Logic rules used in integration with the Reinforcement Learning}
\label{tab:fol_rules}
\begin{center}
\begin{tabular}{ r l }
\toprule
 \texttt{do\_not\_hit\_stone:} & $\forall x, \exists y \quad Agent(x) \land Stone(y) \land AreClose(x,y) \Rightarrow \lnot Move(x,y)$ \\
 \texttt{attack\_enemies:} & $\forall x, \exists y \quad Agent(x) \land Enemy(y) \land AreClose(x,y) \Rightarrow Attack(x,y)$ \\
 \texttt{move\_to\_key:} & $\forall x, \exists y \quad Agent(x) \land Key(y) \land AreClose(x,y) \Rightarrow Move(x,y)$ \\
 \texttt{do\_not\_repeat\_action:} & $\forall x, \exists y \quad Agent(x) \land Action(y) \land LastAction(x,y) \Rightarrow \lnot Perform(x,y)$ \\
\bottomrule
\end{tabular}
\end{center}
\end{table*}

In Figure \ref{fig:rules_integration}, results for the two environments considered are reported. It is clear that, in both cases, the integration of rules led to a increase in the agent's performance.

\subsection{Imitation Learning Approach Results}
The imitation learning approach implemented in LuckyMera is based on the Behavioral Cloning algorithm.
This method was tested using the \texttt{Room-5x5} environment from MiniHack.
It was selected for its low complexity, which guarantees relatively fast training processes and low space occupation to store the dataset.
The model was trained for five epochs, and then the learned policy was evaluated in a different instance of the environment.
In Figure \ref{fig:bc_results} we present a comparison between the performance of a random agent and a trained model. It is clear that the trained agent is always able to solve the task, being also close to the optimal behavior.

\subsection{LuckyMera-v1.0 Agent Baseline}
As a final validation of our implementation, we designed a LuckyMera agent composed of multiple skills (more details in Appendix \ref{app:skills}) and tested against the most challenging \texttt{NetHackChallenge-v0} environment, which represents the complete NetHack game.
This was done to compare our approach with the participants of the NeurIPS 2021 NetHack Challenge \cite{nle_challenge}, which represent the current \emph{state-of-the-art} models tackling the game of NetHack.

Our agent was able to achieve an average score of 1046.96 and a median of 817.
In Figure \ref{fig:results_challenge}, the results of LuckyMera are put against the highest scoring teams from the challenge.
The agent is able to virtually reach the \nth{6} position, overcoming more than 80\% of the competitors.
The challenge ran for 144 days, with 46 participating teams and 631 overall submissions \footnote{\sloppy{The leaderboard is available at \url{https://www.aicrowd.com/challenges/neurips-2021-the-nethack-challenge}}}.

\section{Conclusion} 
LuckyMera is a flexible, modular, extensible and configurable framework to speed up the development of smart AI agents tackling the NetHack game.
It represents a handy tool to implement and test different solutions, going from symbolic approaches to neural networks, in the field of Reinforcement Learning and Imitation Learning, possibly making also use of foundation language models.
It includes a strong baseline agent, capable of achieving good results in the game, and offers the possibility to easily extend its behavior using external modules.
Such modules can be symbolic rules performing a specific action, or neural models trained on a given task.
The architecture also provides the possibility to create automatically labeled datasets, by storing the experiences of the agent in the form of trajectories made by \emph{state-action} pairs. It is possible to specify the elements of the observations to save, including the language representations from the \texttt{nle\_language\_wrapper}.
The trajectories can be used to train neural models via Imitation Learning, or to fine-tune language models to interact with the environment.

\section*{Conflict of interest statement}
\vspace{-2mm}
All authors declare that they have no conflicts of interest.

\vspace{-4mm}
\section*{Acknowledgements}
\vspace{-2mm}
Research partly funded by PNRR - M4C2 - Investimento 1.3, Partenariato Esteso PE00000013 - "FAIR - Future Artificial Intelligence Research" - Spoke 1 "Human-centered AI", funded by the European Commission under the NextGeneration EU programme.

\bibliographystyle{abbrv}
\bibliography{ref.bib}

\begin{thebibliography}{10}

\bibitem{irlsurvey}
S.~Arora and P.~Doshi.
\newblock A survey of inverse reinforcement learning: Challenges, methods and
  progress.
\newblock {\em Artificial Intelligence}, 297:103500, 2021.

\bibitem{dota2_rl}
C.~Berner, G.~Brockman, B.~Chan, V.~Cheung, P.~D{\k{e}}biak, C.~Dennison,
  D.~Farhi, Q.~Fischer, S.~Hashme, C.~Hesse, et~al.
\newblock Dota 2 with large scale deep reinforcement learning.
\newblock {\em arXiv preprint arXiv:1912.06680}, 2019.

\bibitem{octile_distance}
Y.~Bj{\"o}rnsson and K.~Halld{\'o}rsson.
\newblock Improved heuristics for optimal pathfinding on game maps.
\newblock In {\em Proceedings of the AAAI Conference on Artificial Intelligence
  and Interactive Digital Entertainment}, volume~2, pages 9--14, 2006.

\bibitem{openai_gym}
G.~Brockman, V.~Cheung, L.~Pettersson, J.~Schneider, J.~Schulman, J.~Tang, and
  W.~Zaremba.
\newblock Openai gym, 2016.

\bibitem{gym_nethack2}
J.~Campbell and C.~Verbrugge.
\newblock Learning combat in nethack.
\newblock In {\em Proceedings of the AAAI Conference on Artificial Intelligence
  and Interactive Digital Entertainment}, volume~13, pages 16--22, 2017.

\bibitem{nethack_occupancy}
J.~Campbell and C.~Verbrugge.
\newblock Exploration in nethack with secret discovery.
\newblock {\em IEEE Transactions on Games}, 11(4):363--373, 2019.

\bibitem{gym_nethack1}
J.~Campbell and C.~Verbrugge.
\newblock Exploration in nethack with secret discovery.
\newblock {\em IEEE Transactions on Games}, 11(4):363--373, 2019.

\bibitem{deep_blue}
M.~Campbell, A.~J. Hoane~Jr, and F.-h. Hsu.
\newblock Deep blue.
\newblock {\em Artificial intelligence}, 134(1-2):57--83, 2002.

\bibitem{oracle-sage}
A.~Chester, M.~Dann, F.~Zambetta, and J.~Thangarajah.
\newblock Oracle-sage: Planning ahead in graph-based deep reinforcement
  learning.
\newblock In {\em Machine Learning and Knowledge Discovery in Databases:
  European Conference, ECML PKDD 2022, Grenoble, France, September 19--23,
  2022, Proceedings, Part IV}, pages 52--67. Springer, 2023.

\bibitem{dijkstra}
E.~W. Dijkstra.
\newblock A note on two problems in connexion with graphs.
\newblock In {\em Edsger Wybe Dijkstra: His Life, Work, and Legacy}, pages
  287--290. 2022.

\bibitem{impala}
L.~Espeholt, H.~Soyer, R.~Munos, K.~Simonyan, V.~Mnih, T.~Ward, Y.~Doron,
  V.~Firoiu, T.~Harley, I.~Dunning, et~al.
\newblock Impala: Scalable distributed deep-rl with importance weighted
  actor-learner architectures.
\newblock In {\em International conference on machine learning}, pages
  1407--1416. PMLR, 2018.

\bibitem{nle_challenge}
E.~Hambro, S.~Mohanty, D.~Babaev, M.~Byeon, D.~Chakraborty, E.~Grefenstette,
  M.~Jiang, J.~Daejin, A.~Kanervisto, J.~Kim, et~al.
\newblock Insights from the neurips 2021 nethack challenge.
\newblock In {\em NeurIPS 2021 Competitions and Demonstrations Track}, pages
  41--52. PMLR, 2022.

\bibitem{nld}
E.~Hambro, R.~Raileanu, D.~Rothermel, V.~Mella, T.~Rockt{\"a}schel,
  H.~K{\"u}ttler, and N.~Murray.
\newblock Dungeons and data: A large-scale nethack dataset.
\newblock {\em arXiv preprint arXiv:2211.00539}, 2022.

\bibitem{a*}
P.~E. Hart, N.~J. Nilsson, and B.~Raphael.
\newblock A formal basis for the heuristic determination of minimum cost paths.
\newblock {\em IEEE transactions on Systems Science and Cybernetics},
  4(2):100--107, 1968.

\bibitem{e3b}
M.~Henaff, R.~Raileanu, M.~Jiang, and T.~Rockt{\"a}schel.
\newblock Exploration via elliptical episodic bonuses.
\newblock {\em arXiv preprint arXiv:2210.05805}, 2022.

\bibitem{latent_world_models}
S.~Kessler, P.~Mi{\l}o{\'s}, J.~Parker-Holder, and S.~J. Roberts.
\newblock The surprising effectiveness of latent world models for continual
  reinforcement learning.
\newblock {\em arXiv preprint arXiv:2211.15944}, 2022.

\bibitem{torchbeast}
H.~K{\"u}ttler, N.~Nardelli, T.~Lavril, M.~Selvatici, V.~Sivakumar,
  T.~Rockt{\"a}schel, and E.~Grefenstette.
\newblock Torchbeast: A pytorch platform for distributed rl.
\newblock {\em arXiv preprint arXiv:1910.03552}, 2019.

\bibitem{nle}
H.~K{\"u}ttler, N.~Nardelli, A.~Miller, R.~Raileanu, M.~Selvatici,
  E.~Grefenstette, and T.~Rockt{\"a}schel.
\newblock The nethack learning environment.
\newblock {\em Advances in Neural Information Processing Systems},
  33:7671--7684, 2020.

\bibitem{deep_learning}
Y.~LeCun, Y.~Bengio, and G.~Hinton.
\newblock Deep learning.
\newblock {\em nature}, 521(7553):436--444, 2015.

\bibitem{alvinn}
D.~A. Pomerleau.
\newblock Alvinn: An autonomous land vehicle in a neural network.
\newblock {\em Advances in neural information processing systems}, 1, 1988.

\bibitem{cora}
S.~Powers, E.~Xing, E.~Kolve, R.~Mottaghi, and A.~Gupta.
\newblock Cora: Benchmarks, baselines, and metrics as a platform for continual
  reinforcement learning agents.
\newblock In {\em Conference on Lifelong Learning Agents}, pages 705--743.
  PMLR, 2022.

\bibitem{dagger}
S.~Ross, G.~Gordon, and D.~Bagnell.
\newblock A reduction of imitation learning and structured prediction to
  no-regret online learning.
\newblock In {\em Proceedings of the fourteenth international conference on
  artificial intelligence and statistics}, pages 627--635. JMLR Workshop and
  Conference Proceedings, 2011.

\bibitem{minihack}
M.~Samvelyan, R.~Kirk, V.~Kurin, J.~Parker-Holder, M.~Jiang, E.~Hambro,
  F.~Petroni, H.~K{\"u}ttler, E.~Grefenstette, and T.~Rockt{\"a}schel.
\newblock Minihack the planet: A sandbox for open-ended reinforcement learning
  research.
\newblock {\em arXiv preprint arXiv:2109.13202}, 2021.

\bibitem{chess_rl}
J.~Schrittwieser, I.~Antonoglou, T.~Hubert, K.~Simonyan, L.~Sifre, S.~Schmitt,
  A.~Guez, E.~Lockhart, D.~Hassabis, T.~Graepel, et~al.
\newblock Mastering atari, go, chess and shogi by planning with a learned
  model.
\newblock {\em Nature}, 588(7839):604--609, 2020.

\bibitem{go_rl}
D.~Silver, A.~Huang, C.~J. Maddison, A.~Guez, L.~Sifre, G.~Van Den~Driessche,
  J.~Schrittwieser, I.~Antonoglou, V.~Panneershelvam, M.~Lanctot, et~al.
\newblock Mastering the game of go with deep neural networks and tree search.
\newblock {\em nature}, 529(7587):484--489, 2016.

\bibitem{starcraft_rl}
O.~Vinyals, I.~Babuschkin, W.~M. Czarnecki, M.~Mathieu, A.~Dudzik, J.~Chung,
  D.~H. Choi, R.~Powell, T.~Ewalds, P.~Georgiev, et~al.
\newblock Grandmaster level in starcraft ii using multi-agent reinforcement
  learning.
\newblock {\em Nature}, 575(7782):350--354, 2019.

\bibitem{imitation_learning}
B.~Zheng, S.~Verma, J.~Zhou, I.~W. Tsang, and F.~Chen.
\newblock Imitation learning: Progress, taxonomies and challenges.
\newblock {\em IEEE Transactions on Neural Networks and Learning Systems},
  pages 1--16, 2022.

\end{thebibliography}

\clearpage
\appendix

\section{Usage of the Framework} \label{app:usage}
The LuckyMera framework is designed to be highly configurable at runtime.
The configuration is done via a \texttt{config.json} file and through the command-line interface.
\vspace{-2mm}
\subsection{Configuration File}
\vspace{-2mm}
The LuckyMera framework can be configured using a \texttt{json} file. In Figure \ref{fig:config_file} there is an example of a configuration file.
It is used to set the most relevant parameters that condition the agent behavior.
In fact, the first parameter is the priority list of skills the agent can perform; it highly influences the bot's strategy and playstyle, and hence its performance. 
The user can insert into the list any of the implemented skills, or remove some of them, so that it is easy to test new modules and different approaches.

The second parameter allows the user to choose between the \emph{``standard''} mode and the \emph{``fast''} mode.
In standard mode, the entire NetHack interface is printed when an action is performed, in order to study the agent's behavior.
Instead, the fast mode gives just a few details about the bot's current performance, intended to be used for rapid experimentation.
\begin{figure}[H]
\begin{minted}
[frame=lines,
framesep=2mm,
fontsize=\footnotesize]
{json}
{"skill_priority_list": [
"Pray",
"Eat",
"Elbereth",
"Run",
"Break",
"Fight",
"Gold",
"StairsDescend",
"StairsAscend",
"ExploreClosest",
"Horizon",
"Unseen",
"HiddenRoom",
"HiddenCorridor"
],

"fast_mode": "on",
"attempts": "5"
}
\end{minted}
\caption{Sample configuration file for LuckyMera}
\label{fig:config_file}
\end{figure}

Figure \ref{fig:fast_mode} gives an example of the LuckyMera fast mode; it shows both statistics of the entire run, with the mean and median score of all the games played, and information about the current game, with the score obtained so far and the number of turns played.
Lastly, the user can set the number of games the agent has to perform via the \texttt{attempt} parameter.

\vspace{-2mm}
\subsection{Command-line Interface}
\vspace{-2mm}
LuckyMera offers an handy command-line interface, needed to specify the runtime parameters.
By defining these parameters, it is possible to select the mode of use of the framework.
In fact, LuckyMera provides three main options:
\begin{itemize}
    \item \emph{Inference} mode: Use the configuration specified in the \texttt{config} file to play the game, in order to obtain the score of the agent.
    It is also possible to indicate a subset of the observation keys available in NLE, to select the type of information the agent is allowed to use; it is convenient also in training mode;
    \item \emph{Trajectory saving} mode: the experiences of the agent can be saved in the form of \emph{state-action} pairs. These trajectories can be used to train a neural model.
    It is possible to specify which observation keys to save, and the path where to save them.
    The trajectories can also be saved in language mode, using the \texttt{nle\_language\_wrapper};
    \item \emph{Training} mode: the framework can be used to train a neural model.
    In this case, it is necessary to specify the training algorithm, which must be implemented, and the dataset to use.
    It is also possible to indicate other typical training hyperparameters, such as the number of epochs, the learning rate, the batch size, the scheduler gamma and the random number generator seed.
    Lastly, the user can select the option to use a GPU, to perform the training process faster.
\end{itemize}

\begin{figure*}
    \centering
    \includegraphics[width=0.7\linewidth]{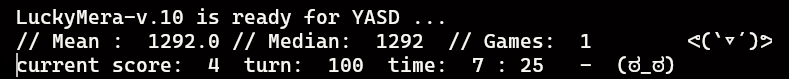}
    \caption{Sample screen of the LuckyMera fast mode}
    \label{fig:fast_mode}


\centering
\captionsetup[table]{labelsep=newline}
\captionof{table}{ Command-line parameters offered by LuckyMera}
\begin{center}
    \begin{tabular}{c|c}
        \toprule
         \textbf{Parameter} & \textbf{Description}  \\
         \midrule
         \texttt{-{}-inference} & Use the framework in inference mode, to actually play the game\\
         \texttt{-{}-training} & Use the framework in training mode. Train a neural model\\
         \texttt{-{}-observation\_keys} & Specify the observation space of NLE\\
         
         \midrule
         \texttt{-{}-create\_dataset} & Use the agent to generate a dataset of trajectories\\
         \texttt{-{}-language\_mode} & Save trajectories in language mode, using the \texttt{nle\_language\_wrapper}\\
         \texttt{-{}-keys\_to\_save} & Specify the observation keys to save\\
         \texttt{-{}-filename} & Path where to save the trajectories\\

         \midrule
         \texttt{-{}-training\_alg} & Select the training algorithm to use\\
         \texttt{-{}-dataset} & Path of the dataset to use for the training process\\
         \texttt{-{}-checkpoint} & Path where to save the trained model\\
         \texttt{-{}-cuda} & Use the GPU for the training process\\
         \texttt{-{}-no\_cuda} & Do not use the GPU for the training process\\
         \texttt{-{}-seed} & Specify the seed for the random number generation\\
         \texttt{-{}-batch\_size} & Batch size for the training process\\
         \texttt{-{}-learning\_rate} & Learning rate for the training process\\
         \texttt{-{}-scheduler\_gamma} & The gamma parameter of the scheduler for the training process\\
         \texttt{-{}-epochs} & Number of epochs to perform during the training process\\
         \bottomrule
    \end{tabular}
\end{center}
\label{table:cli_parameters}
\end{figure*}

In Table \ref{table:cli_parameters} there is the complete list of the available parameters.

\begin{figure*}
\captionsetup[table]{labelsep=newline}
\captionof{table}{ List of the skills already implemented in the LuckyMera Framework.}
    \begin{tabular}{c|l}
        \toprule
        \textbf{Skill Name} & \textbf{Description}  \\
         \midrule
         \texttt{RandomWalk} & It selects randomly a movement action. Used for debugging of the framework\\
         \texttt{NeuralWalk} & Integration of a Reinforcement Learning model for the movement skill\\
         \texttt{BCWalk} & Integration of a Behavioral Cloning model for the movement skill\\
         \texttt{Pray} & In an emergency situation, ask for help to the agent's god\\
         \texttt{Eat} & Reach and eat some food visible in the map to avoid starvation\\
         \texttt{Elbereth} & Engrave the name of the god \emph{Elbereth} to keep monsters from attacking you\\
         \texttt{Run} & Escape from a dangerous situation, if the enemies are too strong for the agent to fight\\
         \texttt{Break} & Take a break to recover hit points. In this situation, the agent performs a search action\\
         \texttt{Fight} & Attack a close enemy, avoiding the agent's pet and passive monsters\\
         \texttt{Gold} & Collect the gold visible on the map \\
         \texttt{StairsDescend} & Search for the position of the descending staircase and go down one level\\
         \texttt{StairsAscend} & Search for the position of the ascending staircase and go up one level\\
         \texttt{ExploreClosest} & Get to the closest door or corridor. Also, opens the door or explores the corridor\\
         \texttt{Horizon} & Reach the furthest visible cells\\
         \texttt{Unseen} & Explore a portion of the map never seen before\\
         \texttt{HiddenRoom} & Search for an hidden room. Hidden rooms can contain helpful items for the agent\\
         \texttt{HiddenCorridor} & Search for an hidden corridor. Hidden corridors can be necessary to continue the game\\
         \bottomrule
    \end{tabular}
\label{table:skills_list}
\end{figure*}

\vspace{-3mm}
\section{LuckyMera Skills} \label{app:skills}
The LuckyMera framework is released with a set of implemented skills; those skills are the one used to obtain the results presented in Section \ref{sec:results}.
The complete list of LuckyMera skills is presented in Table \ref{table:skills_list}.
The list can be modified, by adding or removing skills, and each skill can be easily changed, implementing new approaches.
Further details about the game mechanics can be found at the NetHack Wiki \footnote{\url{https://nethackwiki.com/wiki/Main\_Page}}.

\end{document}